\documentclass[sigconf]{acmart}

\usepackage{booktabs} 

\usepackage{gensymb}

\usepackage{hyperref}

\setcopyright{rightsretained}

\acmDOI{}

\acmISBN{}

\acmConference[DSDAH2018]{KDD }{Workshop on Data Science for Digital Art History}{2018}
\acmYear{2018}
\copyrightyear{2018}

\acmArticle{4}

\begin{document}
\title{Annotating shadows, highlights and faces: the contribution of a `human in the loop' for digital art history.}

\author{Maarten Wijntjes}
\affiliation{%
  \institution{Delft University of Technology}
  \streetaddress{Landbergstraat 15}
  \city{Delft}
  \state{Netherlands}
  \postcode{}
}
\email{m.w.a.wijntjes@tudelft.nl}

\renewcommand{\shortauthors}{M.W.A. Wijntjes}

\begin{abstract}
While automatic computational techniques appear to reveal novel insights in digital art history, a complementary approach seems to get less attention: that of human annotation. We argue and exemplify that a `human in the loop' can reveal insights that may be difficult to detect automatically. Specifically, we focussed on perceptual aspects within pictorial art. Using rather simple annotation tasks (e.g. delineate human lengths, indicate highlights and classify gaze direction) we could both replicate earlier findings and reveal novel insights into pictorial conventions. We found that Canaletto depicted human figures in rather accurate perspective, varied viewpoint elevation between approximately 3 and 9 meters and highly preferred light directions parallel to the projection plane. Furthermore, we found that taking the averaged images of leftward looking faces reveals a woman, and for rightward looking faces showed a male, confirming earlier accounts on lateral gender bias in pictorial art. Lastly, we confirmed and refined the well-known light-from-the-left bias. Together, the annotations, analyses and results exemplify how  human annotation can contribute and complement to technical and digital art history.

\end{abstract}


\maketitle

\section{Introduction}
What is a `picture'? Koenderink \cite{Koenderink2015} often (e.g. \cite{Koenderink2011,Koenderink2012}) refers to the French painter Maurice Denis' famous quote:
\begin{quote}
It is well to remember that a picture before being a battle horse, a nude woman, or some anecdote, is
essentially a flat surface covered with colours assembled in a certain order.
\end{quote}
A picture, Koenderink  \cite{Koenderink2015} continues, implies a doubles sided intentionality: the artist intends that the picture is \emph{looked at}, and not for example used as tea plate, while the observers views the "flat surface covered with colours" as a \emph{picture}, i.e. looks into pictorial space. 

Denis' famous quote can be adapted for digitised artwork as something like \emph{It is well to remember that the \emph{digital documentation of a} picture before being a battle horse, a nude woman, or some anecdote, is essentially a \emph{matrix filled 8 bit colours} assembled in a certain order.} Digital collections of paintings are full of battle horses, nude women and anecdotes while the computational analysis used in some digital art history studies (e.g. \cite{manovich2015data}) focusses on the 'flat surface', the matrix including its features, while leaving out the observer. This omission is not necessarily problematic because machine learning techniques are rather useful for identifying battle horses and nude women. However, understanding `visual art' \cite{Koenderink2015} on a computational level is not limited to categorisations of depicted objects. What differentiates the battle horses of El Greco, Rubens, Degas and Delacroix? It is the 'riddle of style' \cite{gombr}. There are successful attempts in quantifying artistic style computationally \cite{Hughes2010,Johnson2008} but these computations are performed on, say, `generalised' colours/pixels: image features. Also style transfer algorithms \cite{gatys2016image}, although creating wonderful results, rely in essence on image features. 

This paper explores cases where a human hand is (still) needed to extract information from pictures. With a relatively small set of simple image annotations it is possible to generate a substantial amount of relevant insights. We will mainly focus on understanding artists' handling of light in pictorial scenes. Shading an object to reveal its 3D volume, or using cast shadows to denote their position are techniques found in every introduction on art history. Interestingly, these topics are also well represented in the literature about visual perception. The topic of `Shape from shading' \cite{Ramachandran1988}  is concerned with how humans infer 3D shape in the basis of light direction estimates and later was generalised to shape from `x' \cite{Bulthoff}, where `x' can stand for various so-called depth cues like texture gradients \cite{Todd:2004lr}, specular reflections \cite{Fleming:2004cl} and outline curvatures \cite{koend1984}. These `depth cues' are also present in paintings, making the artist a neuroscientist \cite{Cavanagh2005} `avant la lettre'. One may actually wonder how much knowledge about perception is latently present in art history, waiting to be discovered by neuroscientists. 

Besides shading, an important element of pictorial space is the shadow. Whereas the rendering of smooth gradients is seen throughout (western) art history starting around AD (e.g. Roman mosaics, Pompeian wall paintings or Fayum mummy portraits), the rendering of cast shadows appears somewhat more erratic. The most common cast shadow style seems a formless blob `glueing' the object to the ground, a trick important for spatial cognition \cite{mamassian1998perception}. It is not uncommon for painters to neglect the correct projective transformation and simply copy the frontal outline \cite{casati2008}. Interestingly and emphasising the intricate connection between art and perception, human observers are rather unaware that this rendition of shadow is physically impossible \cite{ostrovsky2005perceiving}. Further evidence that visual perception is rather insensitive to perspective errors \cite{Pont2011} makes the \emph{accurate} rendering of cast shadows a rather ungrateful effort. Despite the observers' insensitivity to accuracy, it is impossible to deny the importance role of cast shadows in art history \cite{baxandall1997shadows,gombrich1995shadows}.

Another depth cue (but also a \emph{material} cue) is the highlight. The reflection of the light source on the shiny surface denotes but second order (i.e. curvature) shape geometry \cite{Fleming:2004cl} as well as signalling the level of gloss through features like contrast and sharpness \cite{Pellacini2000}. It makes sense to make a distinction between the rendering of detailed reflections of the environment in metal or glass like objects, and the application of a simple dot, ellipse or line (e.g. on grapes or eyes). Especially, highlights in the eye (we will coin them `eyelights') are an interesting feature because the eyes are locally spherical (actually a double sphere, \cite{Johnson2007}), and can thus be used to reconstruct the 3D light direction. 

What can we do with this information? What do these shadows, highlights etc tell us about the practice of creating images? How does the artist create a picture? These appear questions raised throughout scientific debates (e.g. \cite{hockney2001secret} and \cite{Stork2001}). Besides how a picture was created, these annotations can to some extent capture style and convention. The difference between how two painters depict a similar scene can be called \emph{style}, it is the visual autograph of the artist. Whereas, this topic is rather complex, a good start seems to be describing the works in terms of light, shade and perspective. Attribution reports of art connoisseurs are quite often full of descriptions that involve terms from perception literature, including light and shade. Conventions are partly similar to style, yet less individual and likely more connected with perception \cite{Mamassian2008}. The annotations we will discuss below can all contribute to our understanding of the making process, style and convention.

\section{Annotation cases}

\subsection{Human figures and their shadows}
Besides the pictorial inventions of Giotto and the marvellous rendering of materials by Early Netherlandish artists like van Eyck, it is safe to say that the invention of linear perspective revolutionised the art of painting. While the  geometry of projecting the 3D world onto a 2D surface can be quite tedious, the resulting drawing rules (also known as `secondary geometry' \cite{willats1997art}) are relatively simple and were made readily accessible by Alberti \cite{alberti} around 1435. The main principles describe the relation between straight lines in the world and straight lines on the projection plane (e.g. the panel). In essence: 1) parallel lines in the world \emph{not} parallel to the projection plane converge to a so-called vanishing point, 2) lines on a \emph{plane} in the world converge to a \emph{line} in the pictures (e.g. the horizon is the collection of all lines of the ground plane) and 3) lines in a plane parallel to the projection plane (e.g. the panel) remain undistorted. Although the drawing rules of linear perspective allow for the construction of circular shapes (by first drawing projected squares), the projection of more complex curved shapes is rather difficult. Here, projection tools like the camera obscura may come to help. 

While the projection of architectural structures seems to have been understood rather swiftly after the introduction of linear perspective, the projection of shadows remained a challenge. This seems strange given that cast shadows are caused by parallel light rays from the (very distant) sun and thus follow a simple rule: cast shadows of (say) cylindrical  objects standing upright on a flat ground plane are parallel on the ground. Linear perspective dictates that parallel lines on the ground converge to a vanishing point on the horizon, so the cast shadows should also follow this rule. Although there have been no thorough studies on this (which is partly the motivation behind the current study), our impression is that painters hardly follow this relatively simple rule. 

A particular interesting painter in the context of perspective and cast shadow rendering is Canaletto ($1697-1768$). The cityscapes (knows as `Veduta') offer interesting historical views on Venice among other places. His biographer Zanetti suggested in 1773 that Canaletto made use of a camera obscura, although details about its usage are unknown. Instead of tracing architectural lines, we annotated human figures. The decreasing size of figures closer to the horizon is something that may be easier to accomplish with a camera obscura than through drawing rules. Furthermore, Canaletto's clear rendering of shadows offer a good opportunity to test our hypothesis that perspective of cast shadows may be less accurate then  perspective of objects. 

\subsubsection{Methods}

\begin{figure}[!ht]
\includegraphics[width=0.2\textwidth]{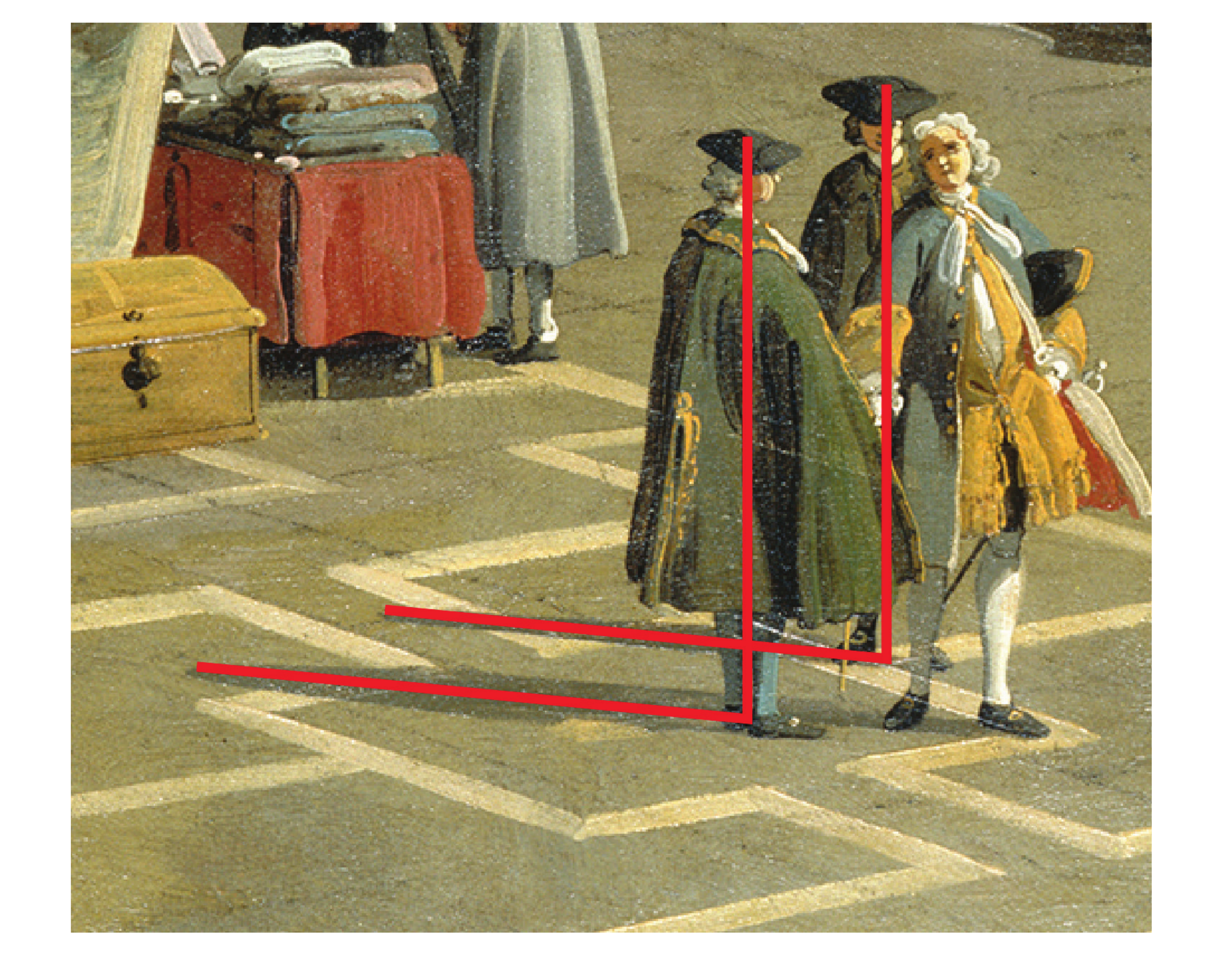}
\caption{The annotation consists of a line from head to toe and (when visible) from toe to shadow head. }
\label{lineAnno}
\end{figure}


We chose 10 paintings by Canaletto and used the presence of people and a flat ground surface together with the availability of a high resolution digital image as selection criteria. The list of paintings can be found in Appendix A. The annotation task consisted of drawing a line from top to bottom of a person, and (if present/visible) continue with a line segment over the cast shadow. Since we were mostly interested in the direction of the shadow, and not so much the length, we also instructed to annotate cast shadows of which the end point was invisible (e.g. because an occlusion). An example can be seen in Figure \ref{lineAnno}. 

\subsubsection{Data analysis}

\begin{figure}[!ht]
\includegraphics[width=0.45\textwidth]{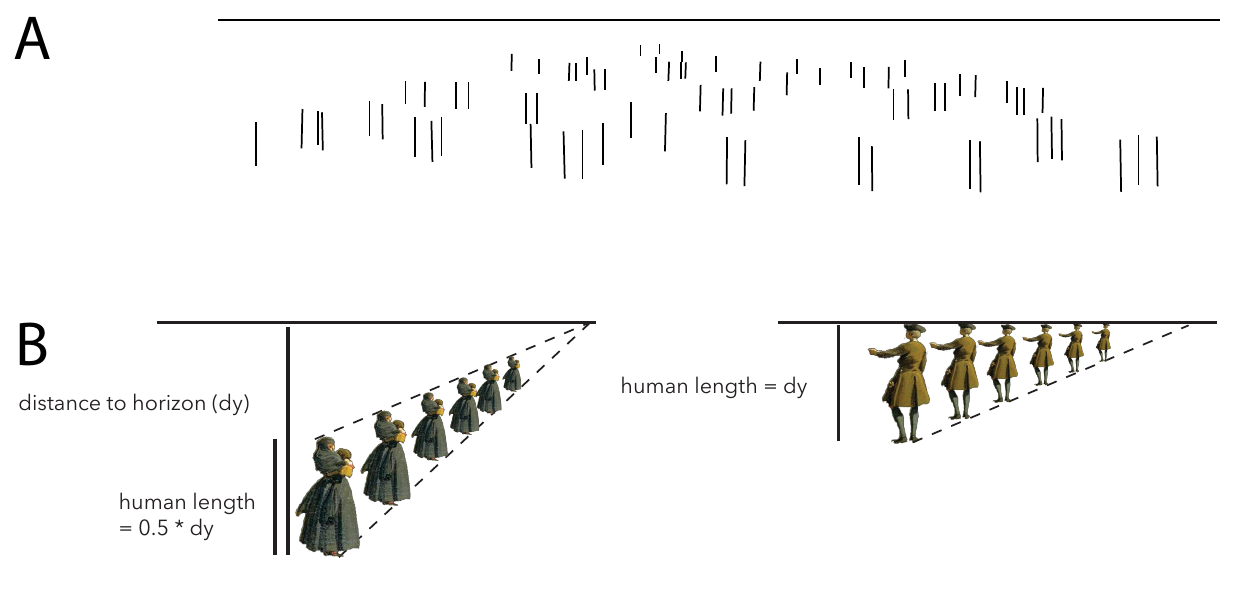}
\caption{A. Example data of annotating position and length of human figures. Note that this rather reduced version of the original still evokes a certain impression of depth. B. Illustration of the meaning of the regression slope: when this slope equals one, the distance between any (!) figure and the horizon should be 1 figure length. When the slope is (for example) 0.5, the distance should be 2 human figures (on the left). Note that we refer to slopes as shown in Figure \ref{sizegradient}.}
\label{viewangle}
\end{figure}

We analysed two aspects: 1) size gradient and 2) shadow convergence. The size gradient (i.e. the gradually decreasing size of objects, in our case the people) was quantified by the length of the first line. We regressed this size on the position with respect to the horizon. This position was denoted by the \emph{feet}, i.e. bottom coordinate of the body line. The regression analysis yields two parameters (offset and slope) and a goodness of fit parameter ($R^2$) that each have a specific meaning. The offset denotes how well the sizes converge to the horizon. We tested whether this offset was significantly different from zero (which would imply that convergence is not at horizon but at a different height). The slope relates to the height of the centre of projection (the location of the painter). A high slope implies a low perspective elevation. Slopes including 95\% confidence intervals were computed to infer whether paintings differed significantly from each-other. Lastly, the coefficient of determination ($R^2$) was computed to infer accuracy of Canaletto's size gradients. 

\begin{figure}[!ht]
\includegraphics[width=0.25\textwidth]{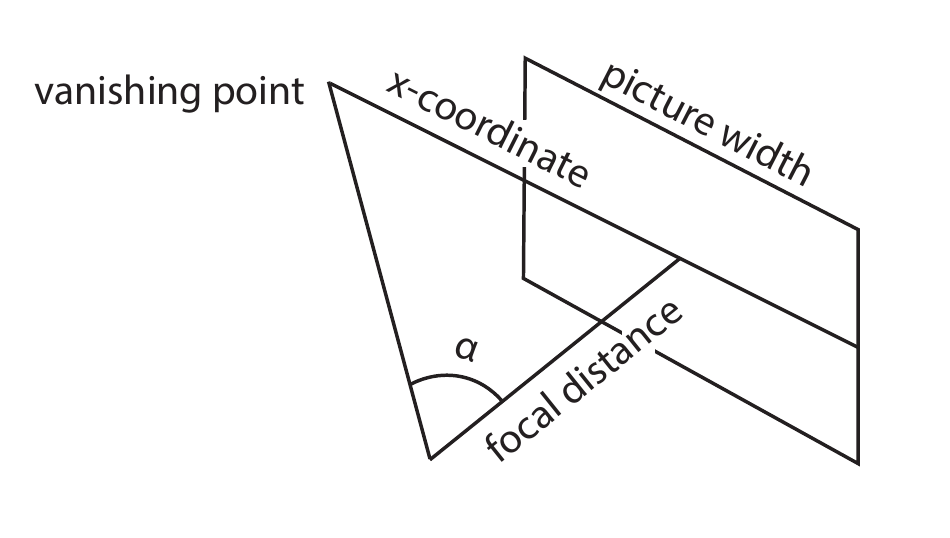}
\caption{To cope with vanishing points at infinity (in cases where the light direction is parallel to the projection plane), the x coordinate was converted into an angle with respect to the (assumed) centre of projection. }
\label{viewangle}
\end{figure}

The shadow convergence was modelled by using the vanishing point on the horizon as variable, while using the RMS of the inner products of model and real (annotated) shadow directions (converted to unit vectors) as cost function. Because the x-coordinate will go to infinity when the shadows are horizontal in the picture plane, this parameter is not very intuitive. Therefore, we converted the vanishing point into an angle, as explained in Figure \ref{viewangle}. The only problem with this parametrisation is that we need to assume a focal distance, which is not readily available and may differ from one picture to another. Nevertheless, we choose to use this parameterisation, and assumed that the focal distance equals the width of the painting. A zero-degree angle means that the vanishing point is in the middle, a 90-degree angle means that shadows are horizontal (in the picture plane) and the vanishing point at infinity.

\subsubsection{Results}
\begin{figure}
\includegraphics[width=0.45\textwidth]{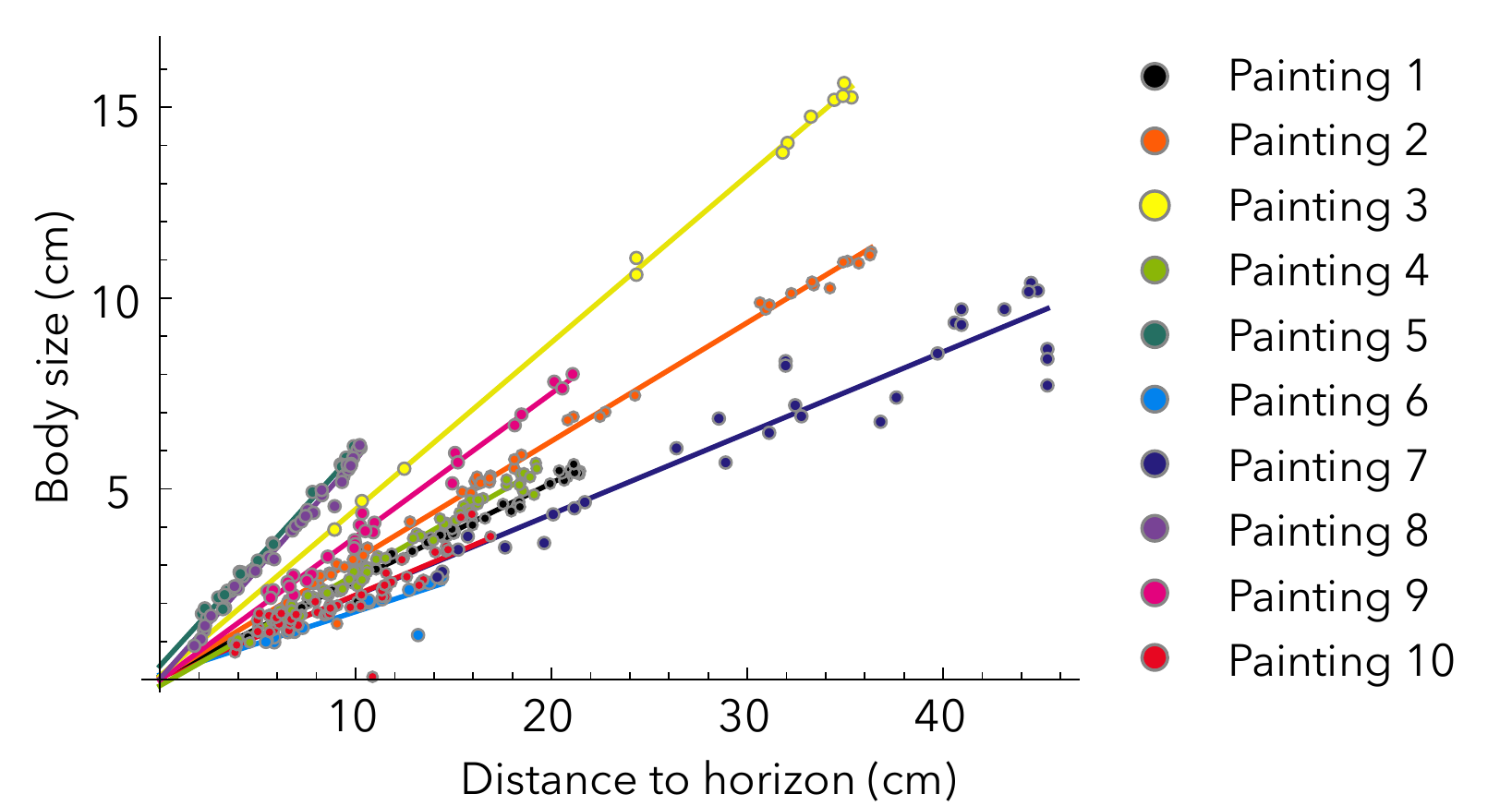}
\caption{Relative sizes of the people depicted in 8 [10?] works by Canaletto.}
\label{sizegradient}
\end{figure}

Firstly, the size gradients were analysed. The human lengths are plotted against their distance to the horizon,  shown in Figure \ref{sizegradient}. The first thing to note is the apparent accuracy with which the human figures have been drawn in perspective. This is confirmed by the $R^2$ values (Table \ref{tabelletje}): all are close to 1 except paintings 6 and 10. This implies that Canaletto quite meticulously drew the sizes of human figures in perspective. Furthermore, the offsets differed significantly from zero only in two cases. This means that Canaletto let the size gradient converge at the horizon level dictated by the architecture (or the actual horizon when visible). Both findings indicate that Canaletto use rather accurate techniques for painting in linear perspective.

\begin{table}
  \caption{Linear regression parameters}
  \label{tabelletje}
  \begin{tabular}{ccll}
    \toprule
    Painting&$R^2$&Offset (p-value) & Slope (95\%CI)\\
    \midrule
    1 & 0.99 & 0.08 (n.s.) & 0.25 (0.25,0.26)\\
    2 & 0.99 & 0.04 (n.s.) & 0.31 (0.3,0.32)\\
    3 & 1.00 & 0.09 (n.s.) & 0.44 (0.43,0.45)\\
    4 & 0.99 & -0.16 (p<0.01) & 0.29 (0.28,0.3)\\
    5 & 0.99 & 0.36 (p<0.01) & 0.56 (0.54,0.58)\\
    6 & 0.78 &  0.12 (n.s.) & 0.17 (0.12,0.21)\\
    7 & 0.90 & 0.08 (n.s.) & 0.21 (0.18,0.24)\\
    8 & 0.99 & 0.02 (n.s.) & 0.57 (0.55,0.6)\\
    9 & 0.99 & -0.02 (n.s.) & 0.38 (0.36,0.39)\\
    10 & 0.77 & 0.05 (n.s.) & 0.22 (0.18,0.25)\\
  \bottomrule
\end{tabular}

\end{table}

\begin{figure}
\includegraphics[width=0.45\textwidth]{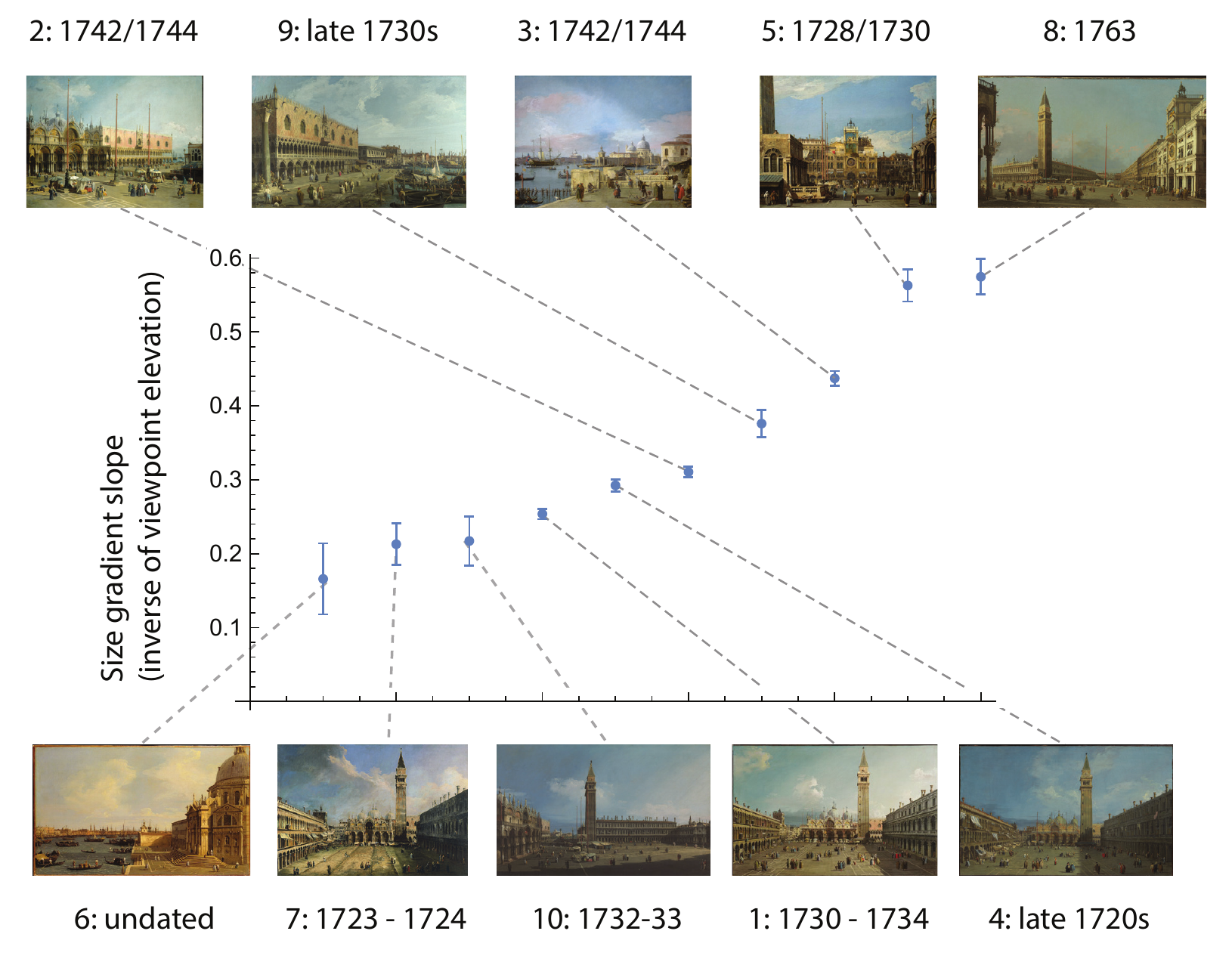}
\caption{Slopes of size gradients including error bars indicating 95\% confidence intervals. The data is sorted from low to high. }
\label{errordingesEdit}
\end{figure}

The relevance of the last parameter is of a different kind. Instead of `accuracy', the slope of the size gradient relates to the height of the viewpoint, the `camera standpoint', which is an element of style.  The regression slopes varied from painting to painting, ranging between 0.17 and 0.57. These boundary values convert to 1.7 and 6 human lengths, respectively. Assuming an average human length of about 1.65cm in the 18th century, this would imply elevation ranging between about 3 and 9 meters. In Figure \ref{errordingesEdit} the slopes and paintings are shown. When viewing these paintings, it indeed appears like the viewpoint starts high (painting 6) and eventually is lowered to close to ground level (painting 8). Interestingly, it appears as if Canaletto preferred a high viewpoint in his early work (1720s and early 1730s), and later descended to a viewpoint closer to the ground. 

\begin{figure}[!ht]
\includegraphics[width=0.45\textwidth]{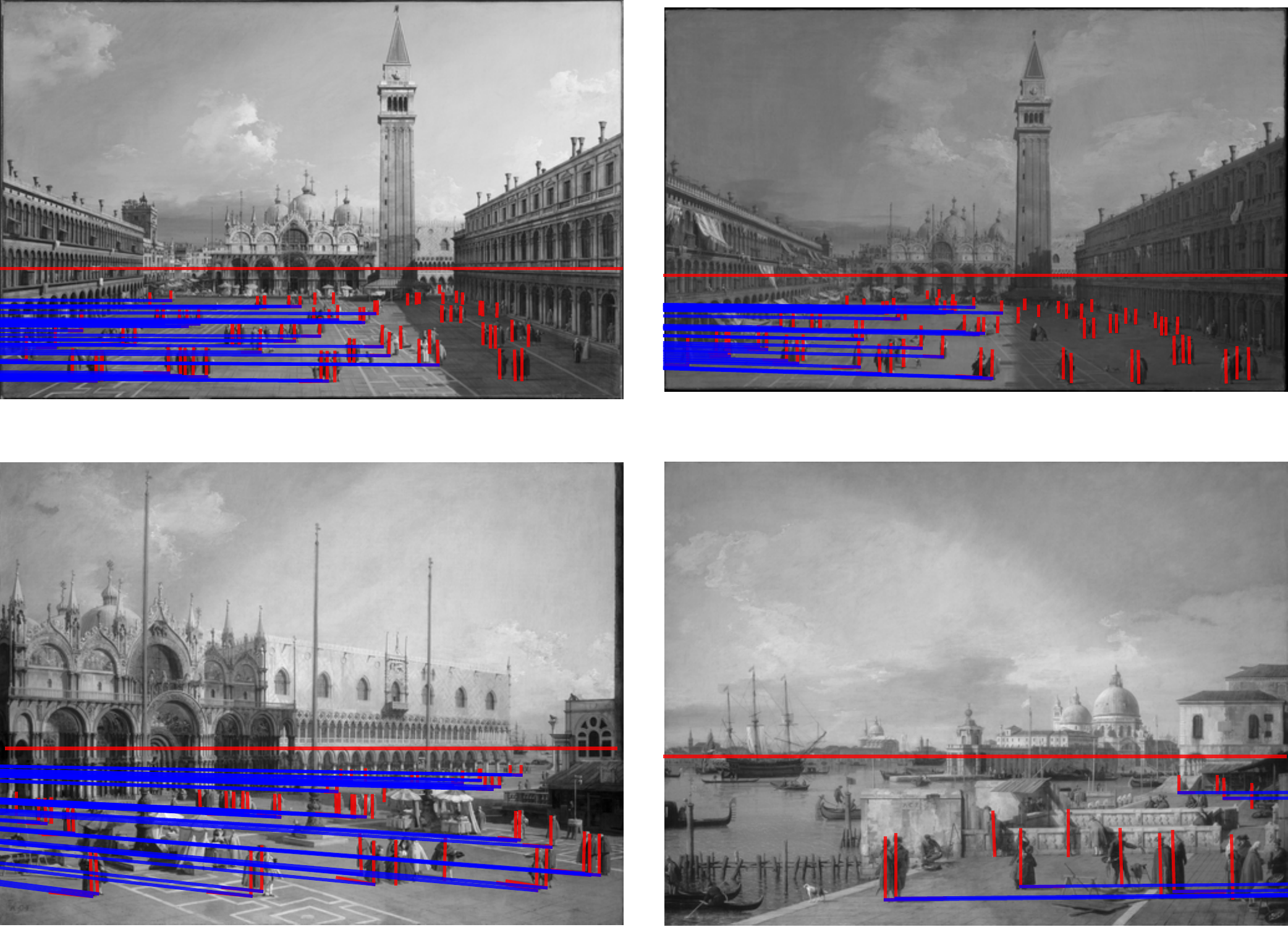}
\caption{Example data and fitted data for 4 paintings. Red lines show human figures and (when annotated) their cast shadow. Blue lines denote the fit.}
\label{exampleData}
\end{figure}

Next to the human figure size gradient, we were interested in how Canaletto paints shadows in perspective. We informally noticed that in many paintings, Canaletto paints shadows horizontal in the picture plane. This implies that in the 3D scene, the light direction should be parallel to the projection plane (and orthogonal to the viewing direction). This particular light direction has the advantage that the shadows are also parallel to the projection plane and thus do not have to converge to a vanishing point. Example data of four paintings is shown in Figure \ref{exampleData}. As can be seen, not all shadows have been annotated. The reason is that Canaletto makes use of opposite shadows when human figures are in an (architectural) shadow. While interesting, these cases can clearly not be captured by our simple model and hence were left out. 

\begin{figure}[!ht]
\includegraphics[width=0.45\textwidth]{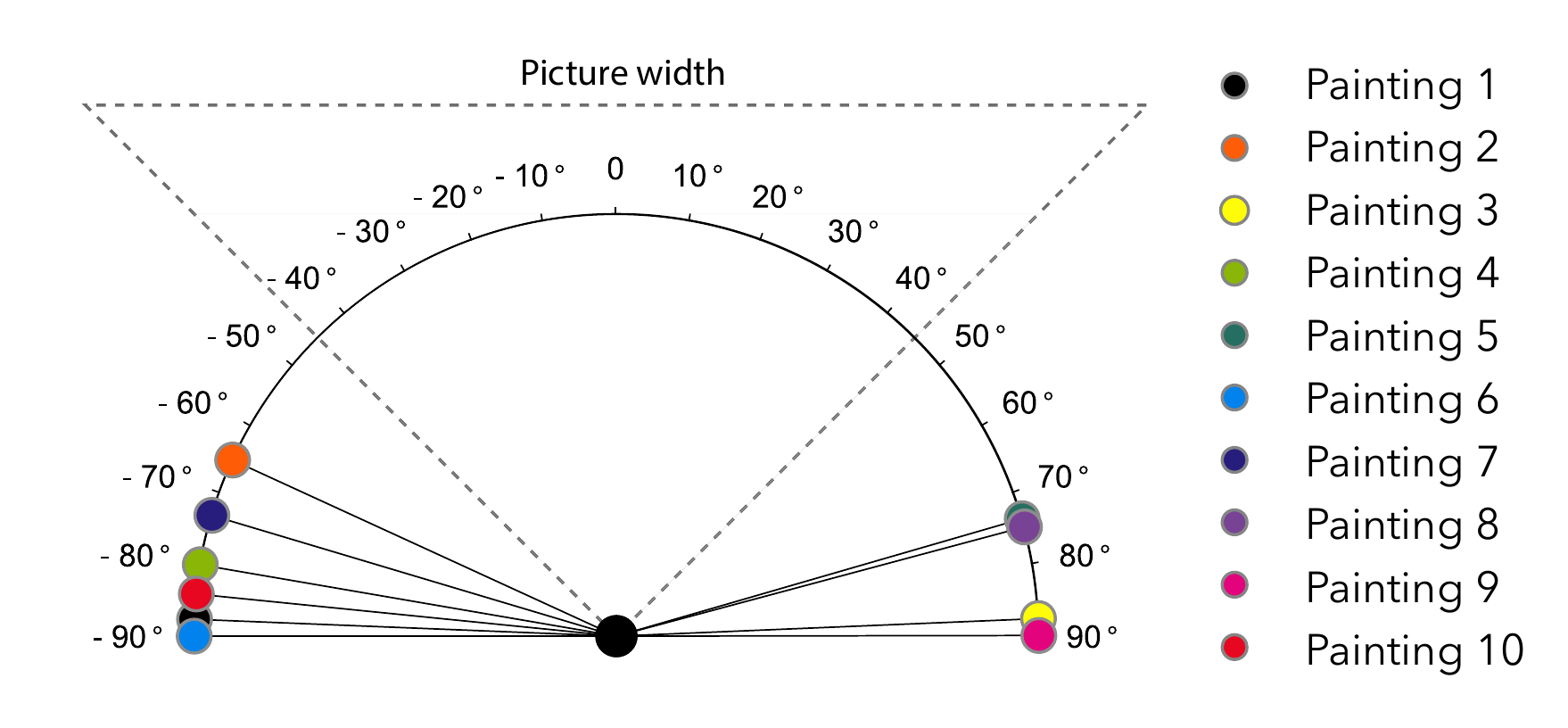}
\caption{Light angles for the 10 paintings. }
\label{shadowFit}
\end{figure}

The fit parameters in terms of light angle (as explained in Figure \ref{viewangle}) are shown in Figure \ref{shadowFit}. As can be seen, in most paintings Canaletto uses light coming either from the exact left, or exact right. Our hypothesis was that in contrast with the meticulous rendering of perspective for the human figures, Canaletto would be less accurate when painting shadows. Concretely, we expected that errors would increase with increasing deviation from parallel lighting. To check, we computed the correlation between the (unsigned) angle and the value of the cost function (as defined in the method section). Indeed, we found a significant correlation ($r=0.78$, $p<0.01$). However, visual inspection of the data suggested that this effect relied on two paintings. Leaving them out made the correlation (r=0.15) insignificant. Therefore, this correlation should be interpreted with substantial restraint. 

When comparing the data in Figure \ref{shadowFit} with the viewpoint analysis in Figure \ref{sizegradient}, it is striking that the paintings with the highest slopes (lowest viewpoint), i.e. paintings 3, 5, 8 and 9, also have a light direction in common. However, we cannot conceive a reason why these two should be correlated.

\subsection{Head pose and gaze directions}

The portrait genre is quite distinct from the Veduta we discussed in the previous section, if only for the light. However, before discussing light annotations in faces, we will discuss head pose and gaze directions. For our analysis on the `eyelights' (section \ref{eyelights}) we needed information about head pose and gaze direction. While collecting and analysing this data we realised that these annotations merit a separate analysis.

\subsubsection{Method}
We used the collection of the Rijksmuseum from which we selected all paintings (N=4190\footnote{this was the number at the moment this study was conduced, at the moment of writing this number approached 5000}). The Viola-Jones algorithm \cite{viola2004robust} implemented in Matlab was used to detect the faces (N=1389). We wanted to label the faces  both with respect to head orientation and gaze direction, e.g. a face directed leftward may look towards the painter, but can also look leftward. There are also cases where the head pose and gaze orientation are opposite (e.g. left gazing while right facing), but we did not analyse these and categorised them as `other'. Also cases where the head pose was not in a horizontal direction were grouped as `other'. In total, 13\% of the faces ended up in the `other' category. 

It should be noted that some statistical patterns may be due to the Viola-Jones algorithm, e.g. it may be biased in detecting frontal faces better than side viewed faces. This should be taken into account when trying to interpret the data. 

\begin{figure}[!ht]
\includegraphics[width=0.45\textwidth]{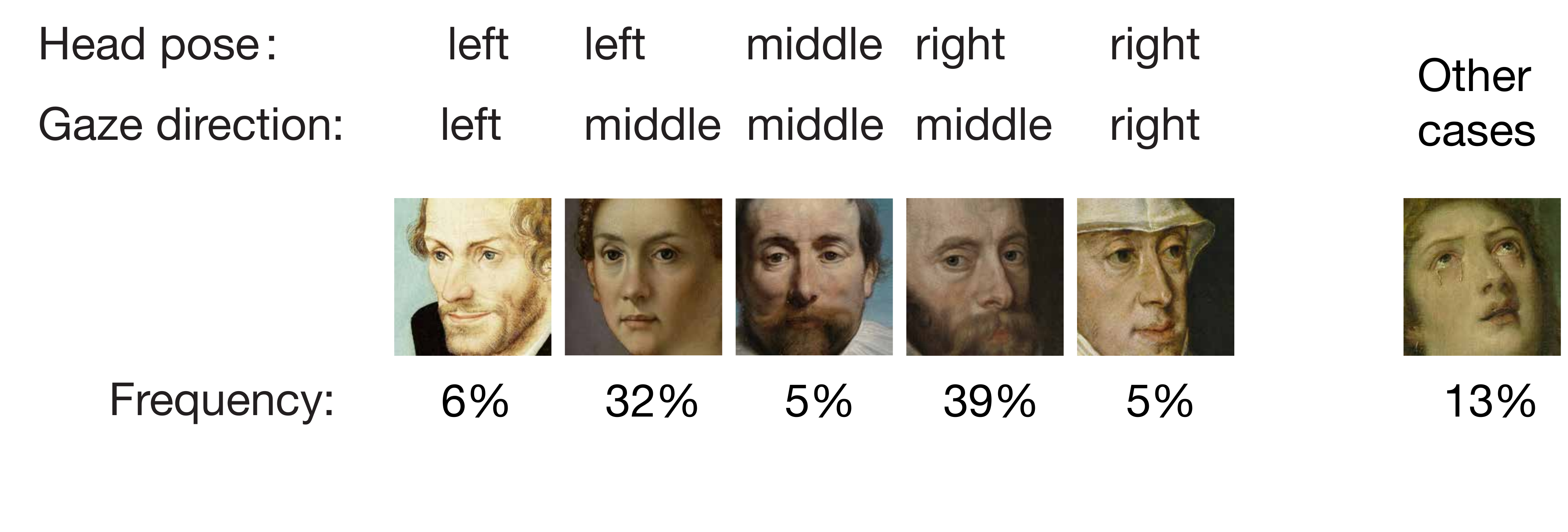}
\caption{Examples of the six categories we labelled, including the mean results. }
\end{figure}

\subsubsection{Results and discussion}
\begin{figure}[!ht]
\includegraphics[width=0.45\textwidth]{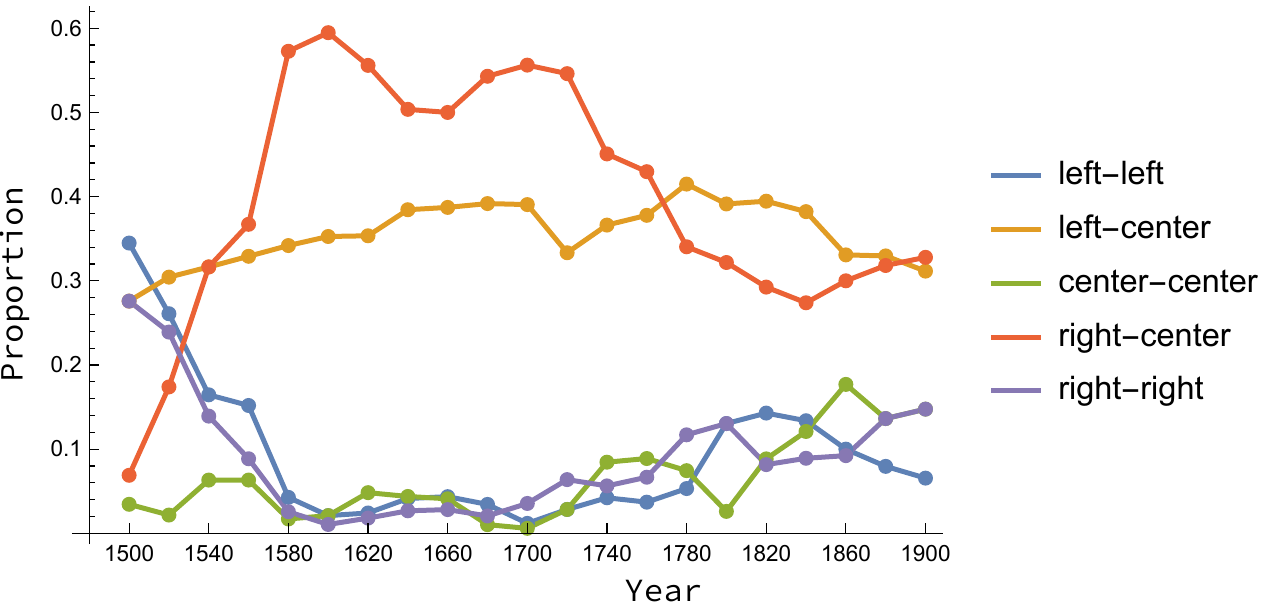}
\caption{Five different gaze/face classes over time. On y-axes is the relative proportion (that should sum up to 1)}
\label{posteryears}
\end{figure}

In Figure \ref{posteryears} the relative proportions of the five different combinations are shown. The first thing to notice is the relatively low presence of sideways oriented faces. As mentioned, this could be due to the algorithm detecting frontally gazing faces better. A finding that is less likely due to algorithmic bias is the relative absence of frontally oriented faces. This is  a well-known phenomenon\cite{McManus1973}.

Lastly, Figure \ref{posteryears} shows that most heads are in a lateral orientation, while gazing frontally. Moreover, there seems a bias towards heads oriented towards the right. It should be noted that a head oriented to the right implies that the viewer sees the right cheek. In literature the laterality terminology is often based on which cheek is shown, so it is convenient that our left-right label (motivated from an annotation perspective) is similar. Interestingly, literature has repeatedly reported a leftward bias \cite{McManus1973,Nicholls1999,Manovich2017}. Here, instead, we find centred around the Dutch Golden Age a rightward bias. There are some possible explanations for this discrepancy between our data and the literature. Firstly, it could be due to something particular in the museum collection (Rijksmuseum). Secondly, we analysed \emph{faces} while most studies analysed \emph{portraits}. 

\begin{figure}[!ht]
\includegraphics[width=0.48\textwidth]{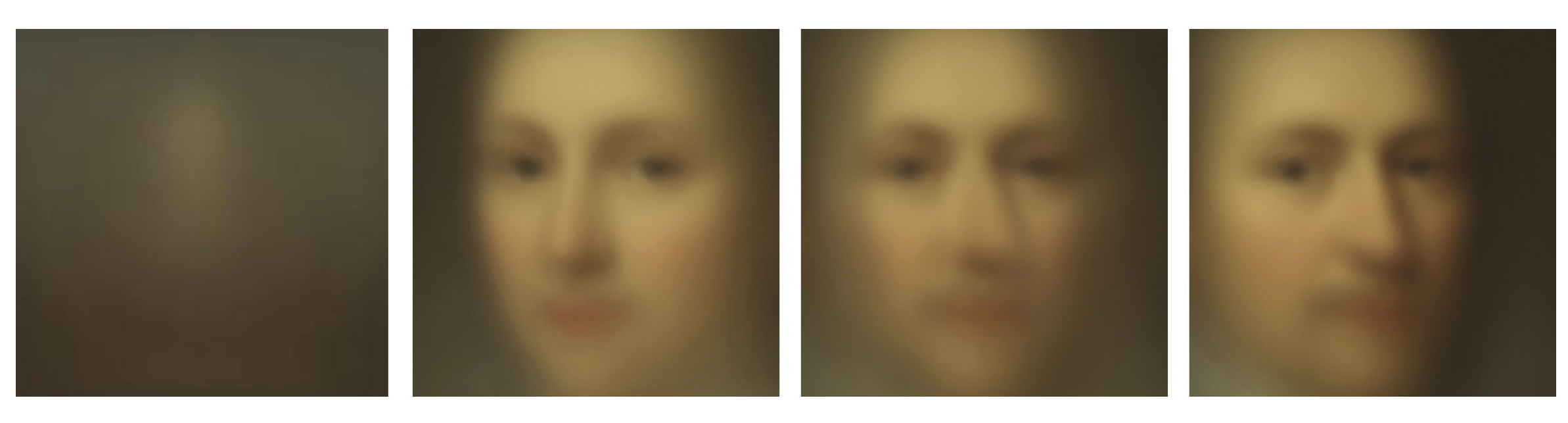}
\caption{Averaged images. On the left, all paintings from the Rijksmuseum: there is some prototypical portrait composition visible, but for the rest it appears uninteresting. However, for faces an interesting effect can be observed: women generally look to the left, while males to the right.}
\label{meanfaces}
\end{figure}

An interesting way of exploratory analysis of multiple images is taking the average \cite{Torralba2003}. When this is done on an unselected set of images (in this case a collection of 4190 paintings), the results are as unspecific as the input is (left side of Figure \ref{meanfaces}). However, the more interesting structures may emerge when averages are taken over a \emph{selection}, in our case over all faces. We first took the average over all faces, displayed right of the middle in Figure \ref{meanfaces}. Then we split the faces in the two laterally labelled directions and found a surprisingly convincing rendering of a female (facing left) and male (facing right). In contrast with the discrepancy between our finding of right cheek dominance with left cheek dominance in literature, this gender difference if much in line with previous research \cite{McManus1973,Nicholls1999}. 

While the relation between gender and head pose is well known, the way we replicated this finding is novel. Torralba \cite{Torralba2003} used averaged images to demonstrate that certain objects naturally reside in certain environments. Making use of annotated object labels it is possible to overlay and average all images with the annotated object in the middle. In cases where the annotations are inaccurate or absent, specialised algorithms \cite{Zhu2014} may help. Using averaged images in art history thus needs spatial annotations, the average of all paintings merely gives a trivial result. Our case of faces is obviously special because of the availability of face detection algorithms. In other cases, such as the painting of grapes or buildings, detailed annotation would be needed.

\subsection{Eyelights}\label{eyelights}
It is well known from the literature on visual perception that the visual system assumes that light comes from above. These types of assumptions are often necessary because the computational problem of vision (constructing a 3D world from a 2D retinal image) is `ill-posed' \cite{bertero1988ill} and confounded \cite{barrow1978recovering}. A famous example is described by Ramachandran \cite{Ramachandran1988}: a disk with light-dark gradient from top to bottom is interpreted as a (3D) convex region, while the opposite gradient cause a concave interpretation.

It was subsequently found that the preferred light direction is more delicate than simple `above': there was also a slight preference for light coming from the left. While the light-from-above assumption is clearly aligned with the statistics of our natural environment (including indoor lighting), the lateral bias is difficult to explain. The authors \cite{sun1998} start their discussion of possible explanation by referring to the famous Gestalt psychologist Metzger: using desk-lamps at the left side does not create shadows for right hand writers. Thus, humans may get conditioned by this reoccurring exposure of light coming from the left and therefore develop the lateral preference. Indeed, \cite{sun1998} found a significant correlation between handedness and preferred light direction. They also tested preferred light direction in paintings and found the same bias. Interestingly, the authors hypothesised that the \emph{cause} of this depiction bias lies in something ``higher perceptual'', while it is quite well known that the depiction bias is actually similar to Metzgers' idea: portrait painters using their right hand while positioning the sitter right of the window so the pencil does casts a (harmless) shadow behind the painters' hand. An alternative hypothesis would be the opposite: that the depiction bias \emph{causes} the perceptual bias, instead of the other way around. This seems plausible since the light prior (as it is called in literature) can be modified by experience \cite{Ada2004}. The influence of the pictorial world on the perception of the real world seems a very attractive topic, although up to now there is little activity in this area. 

Light direction seems thus a rather interesting topic to study further, and the contribution of the work presented in this section is the use of a rather unambiguous annotation method and some ideas on how to analyse them. Instead of letting observers look at the whole face and estimate the light direction using a protractor \cite{sun1998}, we simply annotate the position of the pupil and highlight. Although interesting by itself, the protractor method may induce human bias, i.e. it involves \emph{perception}. How humans estimate light direction from a certain illuminated object (in this case a face) is part of a larger field of research (e.g. \cite{Koenderink2007a,Pont:2007vz,Shea2010}) which makes it both interesting but also complex. Annotating the position of highlights and pupils seems to be prone less to human bias. Nevertheless, it should be noted that a painting is not a photograph and painters may shade the skin differently than would be expected on the basis of a highlight direction.

\subsubsection{Methods}

\begin{figure}[!ht]
\includegraphics[width=0.4\textwidth]{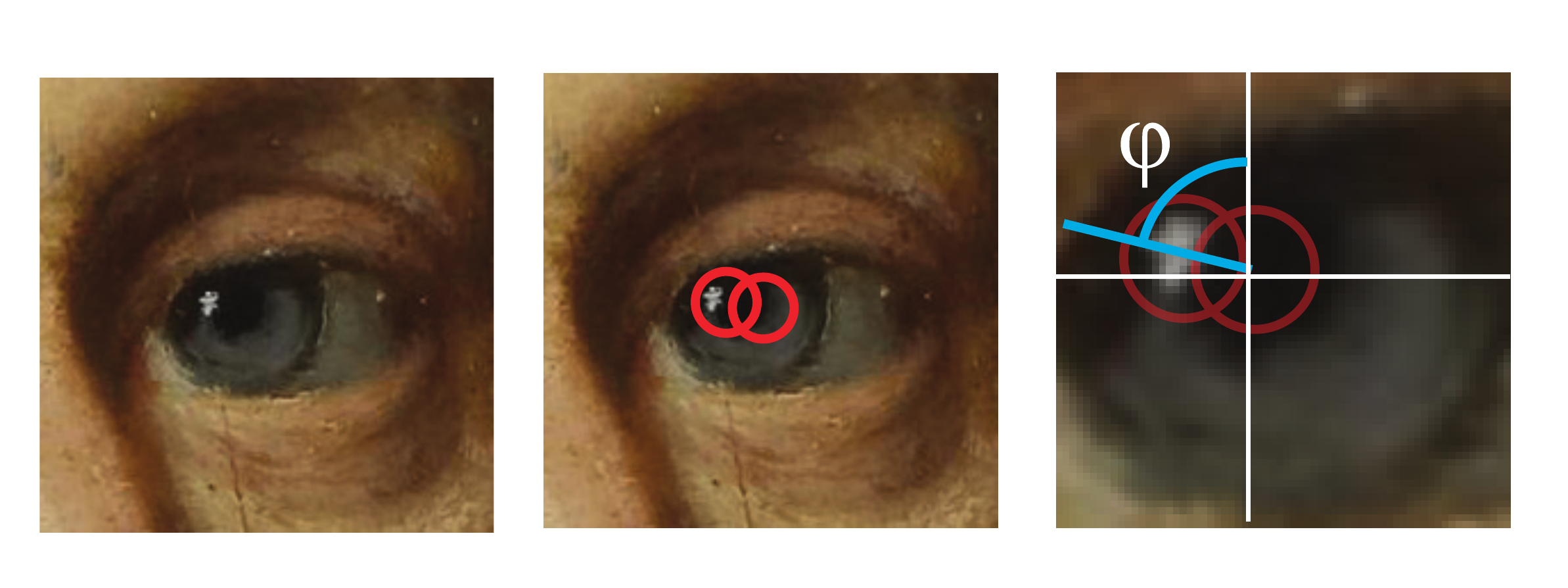}
\caption{ }
\label{eyeAnnotation}
\end{figure}

We used a subsample (N=353) of the 1389 faces while ensuring that the number of faces per time interval stayed constant. Only faces that were laterally oriented, but gazing frontally (towards the viewer) were selected. The annotation consisted of positioning a red circle in the centre of the pupil and in the centre of the highlight. The size of the red circle was constant. We annotated both eyes. As can be seen in Figure \ref{eyeAnnotation}, we converted the positions of pupil and highlight into a single parameter: the tilt angle. It is theoretically possible to also extract the slant angle, but then we also needed the size of the iris which we did not annotate. 

\subsubsection{Analysis}
\begin{figure}[!ht]
\includegraphics[width=0.4\textwidth]{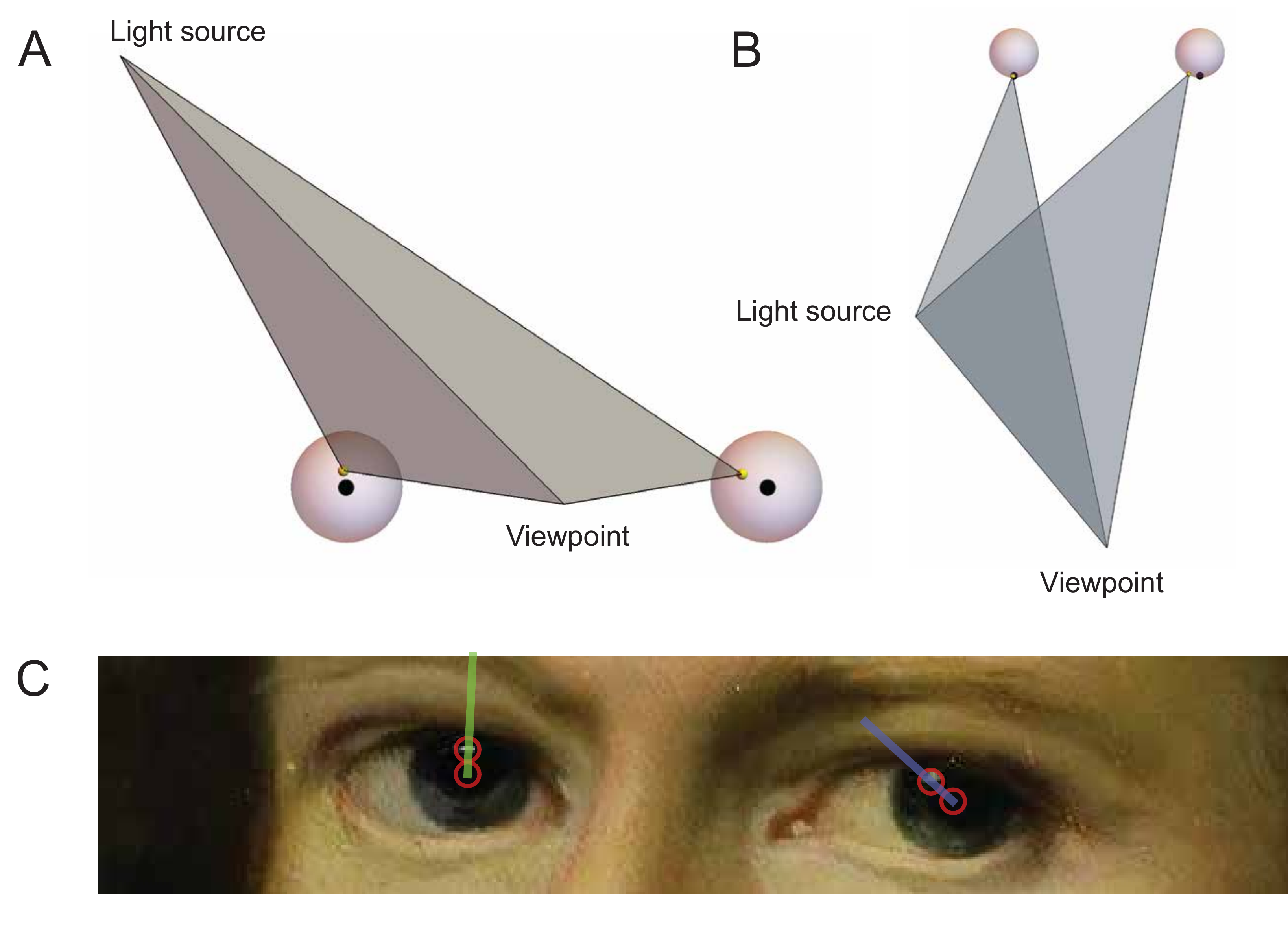}
\caption{A. 3D model of the geometry of highlights. The viewpoint is frontal, towards the eyes. In the top left, a point light source illuminates the eyes and rays causing the highlights are shown. As can be seen, the highlight position (yellow ball) on the left eye is different from the right eye. B. Same as A but now a top view. C. Example from an actual portrait, including the highlight vectors used for the analysis. }
\label{eyepersp}
\end{figure}

The tilt angle was the main parameter that was analysed. However, another interesting aspect can be quantified when the highlight angles are known. Theoretically, when the light source is at a finite distance from the eye, the eyelights should be at different relative positions. For example, if light comes exactly from above (and is close), the highlight in the left eye should be to the right and vice versa.  The geometry and an example are shown in Figure \ref{eyepersp}. Without going too much into geometric detail, we can safely state that for cases where the light comes from above, the angle in the left eye should be smaller than the right eye (in the parametrisation as shown in Figure \ref{eyeAnnotation}), and for light coming from below the opposite should hold. 

\subsubsection{Results}
\begin{figure}[!ht]
\includegraphics[width=0.4\textwidth]{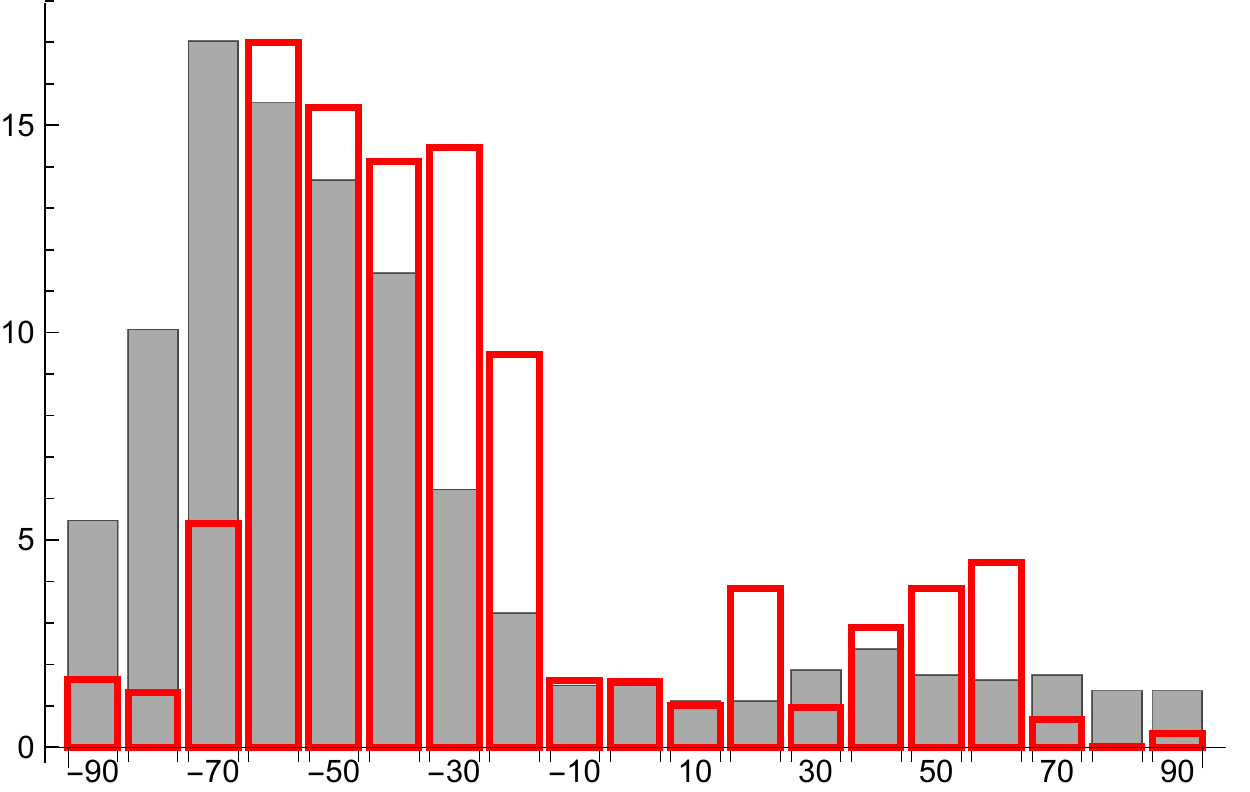}
\caption{Our data (grey bars) and data from Sun and Perona (red outlines). On the y-axis, the relative percentage is shown.}
\label{ourAndPeronaDataPercentages}
\end{figure}

We first created a histogram over all periods. In Figure \ref{ourAndPeronaDataPercentages} we superimposed the data of \cite{sun1998} over our data. Since we used an opposite angle definition, we inverted all data (i.e. 45 degree becomes -45 degrees). As is clearly visible: both histograms look remarkably similar. This is remarkable because both the collections and the annotation methods were rather different. Therefore, light direction seems a very robust convention in art history. 

One difference appears to be the overall mean. In our case, the mean light direction is $-41.1 \degree$ while for \cite{sun1998}  it is $-28.6 \degree$. Thus, our light directions are more extreme. Whether this is due to the annotation method (including human bias), the artists or something else cannot be concluded from the current data, but seems worthwhile to investigate further.

\begin{figure}[!ht]
\includegraphics[width=0.4\textwidth]{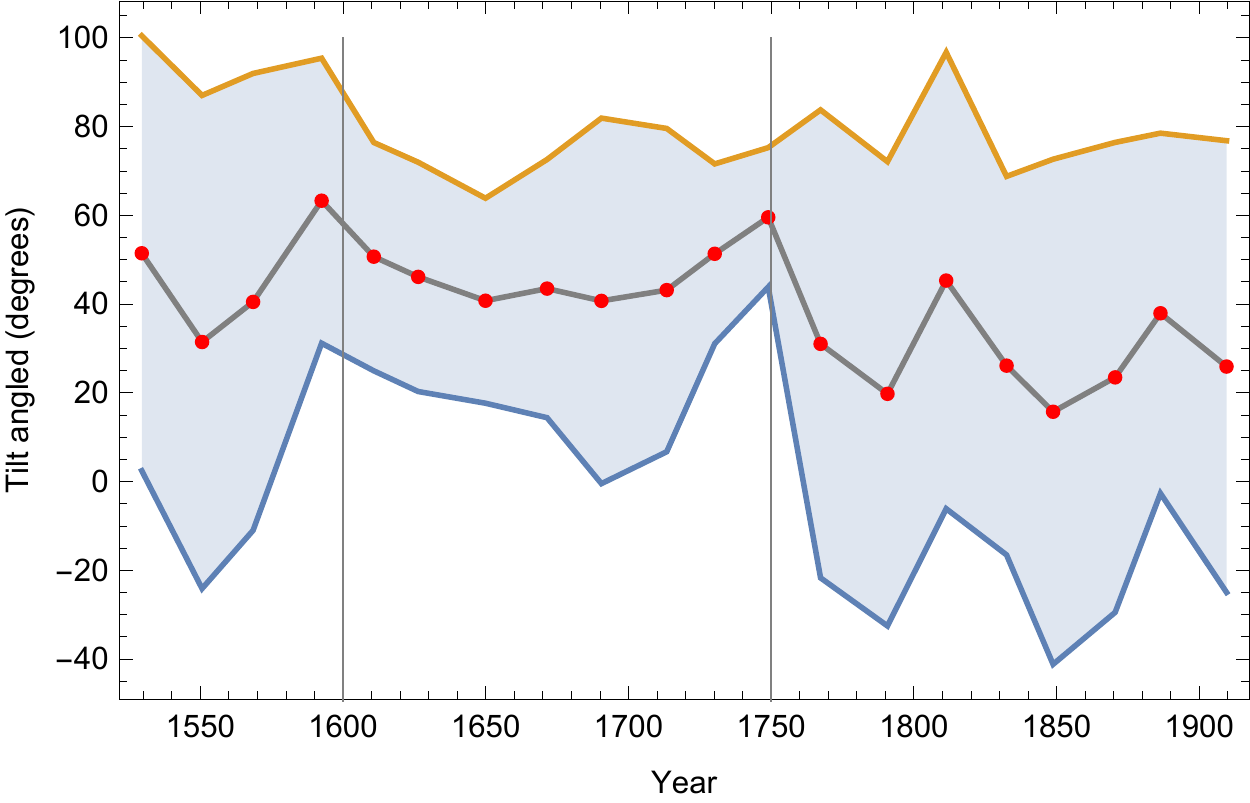}
\caption{Mean and standard deviation for the tilt angle}
\label{eyeOeverTime}
\end{figure}

Next, we analysed whether light direction would change over time. The mean data is plotted in Figure \ref{eyeOeverTime} together with standard deviations. The first thing to notice is the large spread of the data. Almost all mean data fall within the range of the standard deviation edges. Thus, it is relatively difficult to interpret this data. Although speculation, it appears there is a smooth pattern visible between 1600 (start of the Dutch Golden Age) and 1750.

\begin{figure}[!ht]
\includegraphics[width=0.45\textwidth]{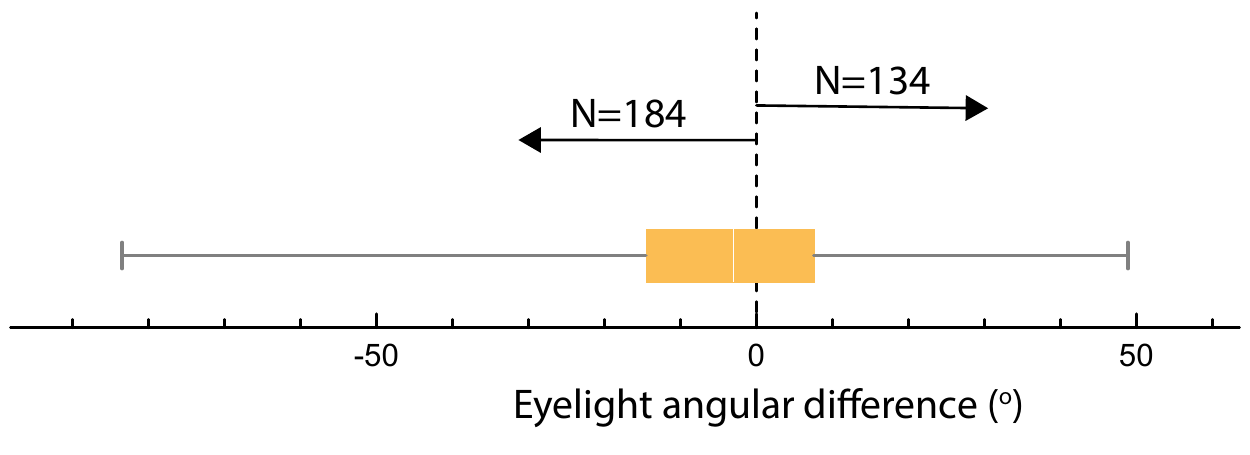}
\caption{Interocular angular highlight difference plotted in a box-whisker plot denoting the quartiles (white line denotes the median)}
\label{boxwhiskerEditted}
\end{figure}

Lastly, we analysed the interocular eyelight difference. For each pair of eyes, we calculated the angular difference between the left and right eye. We only used cases where the light came from above (the large majority). A negative value of this difference  (right angle larger than left angle) implies that the painter takes into account the perspective of highlights as explained in Figure \ref{eyepersp}. In Figure \ref{boxwhiskerEditted} the data is plotted. As can be seen, the majority of angular differences is negative (184 vs 134). This was confirmed by a t-test ($t(317)=-3.197$, $p=0.002$).

\section{Conclusion}
We have attempted to demonstrate the value of human annotations in the analysis of digitised paintings. First, we will summarise the contributions of this paper and then we will discuss these findings in the context of digital art history. The contributions are split in three levels:  annotation, analysis and results, because for all three levels are to some extent independent: one annotation does not necessarily demand a specific analysis, and the same result can be accomplished through different annotations.  \\

\begin{samepage}
\noindent
\textbf{Annotation technique}
\begin{itemize}
\item With a simple \emph{line element}, the height of depicted human figures can be computed. The line element can also be used to annotate cast shadow direction. 
\item To categorise head pose and gaze direction we used \emph{example images}. 
\item For light direction we annotated the \emph{locations of pupil and highlight}. 
\end{itemize}
\end{samepage}

\begin{samepage}
\noindent
\textbf{Annotation analysis}
\begin{itemize}
\item The slope of the size gradient regression denotes the reciprocal of \emph{viewpoint height}. 
\item This vanishing point can be (intuitively) represented by a \emph{vanishing angle}, which denotes the actual 3D light direction if the painting was a window. 
\item Using \emph{averaged images} (in our case faces) can serve as an interesting exploratory tool to reveal common structure within specific motifs. 
\item In contrast with general light direction annotation,  \emph{pairwise eyelight annotations} reveal light perspective handling. 
\end{itemize}
\end{samepage}

\begin{samepage}
\noindent
\textbf{Empirical findings}
\begin{itemize}
\item Canaletto painted human figures in perspective with remarkable accuracy.
\item Based on the size gradient slopes, Canaletto's viewpoint varied between 1.7 and 6 human lengths. 
\item For the paintings we analysed, Canaletto primarily used light directions parallel to the projection plane. 
\item Earlier findings of the rareness of frontal face pose were replicated.
\item Earlier findings of left cheek dominance were contradicted: rather we found right cheek dominance but also variations in time. 
\item Earlier findings on  head pose gender bias were confirmed.
\item Light from the left was confirmed although with different mean values then previous findings. 
\item Converging eyelights indicate  awareness of highlight perspective. 
\end{itemize}
\end{samepage}

\vspace{0.5cm}

Why a `human in the loop for digital art history'? Is it not possible to use contemporary neural networks to detect human figures, faces and highlights? It probably is possible, for faces there exists state of the art algorithms that can compute head pose in three directions while at the same time including age and gender. However, computing gaze direction may be more difficult. While the computer may catch up and also be able to estimate gaze direction, there could always be other aspects to annotate for which algorithms are not ready. Therefore, we think it is a rather valuable and complementary approach next to the development of fully automatic analysis of art history. 

Another reason to make use of human annotations is that it could be more accessible for scientists with little computational background. Although programming (simple) visual interfaces, data collection and analysis should not be underestimated, at least the researcher does not need to have a degree in computer science. Again, this may change in the future and computational analysis may become better accessible. Yet, letting a class of art history students perform an annotation task could be rather low-threshold and possible even have an educational advantage. 

The `human in the loop' is also the human for which the artwork was made. The art or painting is intricately related to visual perception \cite{Cavanagh2005,Melcher2011}. As a scientific field, visual perception offers a wide variety of topics related to pictorial art (e.g. depth, shape, colour, facial expression, material rendering etc). Vice versa, artists have likely made discoveries that yet have to reach vision science.  

What can we do with knowledge about what light direction, head pose, and viewpoint elevation? These features are part of the \emph{style} of the artwork, together with many other features. As discussed in the introduction, these are elements of style and convention. We confirmed known conventions and found some new ones. Primarily, we wanted to demonstrate that revealing these fundamental aspects of visual representation are merely a few annotations away.

\appendix

\section{List of Canaletto paintings}
\scriptsize
\textbf{Painting 1:}
Title: Piazza San Marco, Venice;
Date: c. 1730 - 1734;
Physical Dimensions: 76.2 x 118.8 cm;
Collection: Harvard Art Museums/Fogg Museum;
Link: \url{https://www.harvardartmuseums.org/collections/object/304349}\\
\textbf{Painting 2:}
Title: The Square of Saint Mark's, Venice;
Date: 1742/1744;
Physical Dimensions: 114.6 x 153 cm;
Collection: National Gallery of Art U.S.A. (Washington);
Link:	\url{https://www.nga.gov/collection/art-object-page.32588.html}\\
\textbf{Painting 3:}
Title: Entrance to the Grand Canal from the Molo, Venice;
Date: 1742/1744;
Physical Dimensions: 114.5 x 153.5 cm;
Collection: National Gallery of Art U.S.A. (Washington);
Link: \url{https://www.nga.gov/collection/art-object-page.32589.html}\\
\textbf{Painting 4:}
Title: Piazza San Marco;
Date: late 1720s;
Physical Dimensions: 68.6 x 112.4 cm;
Collection: Metropolitan;
Link: \url{https://www.metmuseum.org/art/collection/search/435839}\\
\textbf{Painting 5:}
Title: The Clock Tower in the Piazza San Marco;
Date: 1728/1730;
Physical Dimensions: 52.07 x 69.52 cm;
Collection: The Nelson ;
Link:  \url{http://art.nelson-atkins.org/objects/12179/} \\
\textbf{Painting 6:}
Title: Venice: Santa Maria della Salute;
Date: undated;
Physical Dimensions: 47.6 x 79.4 cm;
Collection: Metropolitan;
\url{https://www.metmuseum.org/art/collection/search/435840}\\
\textbf{Painting 7:}
Title: The Piazza San Marco in Venice;
Date: 1723 - 1724;
Physical Dimensions: 141.5 x 204.5 cm;
Collection: Museo Nacional Thyssen-Bornemisza, Madrid;
Link: \url{https://www.museothyssen.org/en/collection/artists/canaletto/piazza-san-marco-venice}\\
\textbf{Painting 8:}
Title: Piazza San Marco Looking South and West;
Date: 1763;
56.52 ? 102.87 cm;
Collection: Los Angeles County Museum of Art;
Link: \url{https://collections.lacma.org/node/247295}\\
\textbf{Painting 9:}
Title: Venice: The Doge's Palace and the Riva degli Schiavoni;
Date: late 1730s;
Dimensions: 61.3 x 99.8 cm;
Link: \url{https://www.nationalgallery.org.uk/paintings/canaletto-venice-the-doges-palace-and-the-riva-degli-schiavoni}\\
\textbf{Painting 10:}
Title: Piazza San Marco, Venice;
Date: c. 1732-33;
Physical Dimensions: 61 x 96.5 cm;
Link:  \url{http://www.fujibi.or.jp/en/our-collection/profile-of-works.html?work_id=1239};

\begin{acks}
This work was funded by the Netherlands Organisation of Scientific Research (NWO). 

\end{acks}

\bibliographystyle{ACM-Reference-Format}
\bibliography{KDD_Wijntjes}

\end{document}